\documentclass[english]{acm_proc_article-sp}
\usepackage[T1]{fontenc}
\usepackage[latin9]{inputenc}
\usepackage{array}
\usepackage{booktabs}
\usepackage{graphicx}

\newcommand{\noun}[1]{\textsc{#1}}
\providecommand{\tabularnewline}{\\}

\def\sharedaffiliation{%
\end{tabular}
\begin{tabular}{c}}

\begin{document}
\title{Learning DTW Global Constraint for\\Time Series Classification}

\numberofauthors{2}

\author{\alignauthor Vit Niennattrakul\\
 \email{g49vnn@cp.eng.chula.ac.th} 
\alignauthor Chotirat Ann Ratanamahatana\\ \email{ann@cp.eng.chula.ac.th}
\sharedaffiliation \affaddr{Department of Computer Engineering,
Chulalongkorn University}\\
 \affaddr{Phayathai Rd, Pathumwan, Bangkok 10330 Thailand}\\
 \affaddr{+66 (0)2 218 6956}}
\maketitle
\begin{abstract}
1-Nearest Neighbor with the Dynamic Time Warping (DTW) distance is
one of the most effective classifiers on time series domain. Since
the global constraint has been introduced in speech community, many
global constraint models have been proposed including Sakoe-Chiba
(S-C) band, Itakura Parallelogram, and Ratanamahatana-Keogh (R-K)
band. The R-K band is a general global constraint model that can represent
any global constraints with arbitrary shape and size effectively.
However, we need a good learning algorithm to discover the most suitable
set of R-K bands, and the current R-K band learning algorithm still
suffers from an \textquoteleft{}overfitting\textquoteright{} phenomenon.
In this paper, we propose two new learning algorithms, i.e., band
boundary extraction algorithm and iterative learning algorithm. The
band boundary extraction is calculated from the bound of all possible
warping paths in each class, and the iterative learning is adjusted
from the original R-K band learning. We also use a Silhouette index,
a well-known clustering validation technique, as a heuristic function,
and the lower bound function, LB\_Keogh, to enhance the prediction
speed. Twenty datasets, from the Workshop and Challenge on Time Series
Classification, held in conjunction of the SIGKDD 2007, are used to
evaluate our approach.
\end{abstract}
\category{H.2.8}{Database Management}{data mining}

\terms{Algorithms, Performance, Experimentation}

\keywords{Time Series, Classification, Dynamic Time Warping}

\pagebreak

\section{Introduction}

Classification problem is one of the most important tasks in time
series data mining. A well-known 1-Nearest Neighbor (1-NN) with Dynamic
Time Warping (DTW) distance is one of the best classifier to classify
time series data, among other approaches, such as Support Vector Machine
(SVM) \cite{WuC04}, Artificial Neural Network (ANN) \cite{766918},
and Decision Tree \cite{RodriguezA04}. 

For the 1-NN classification, selecting an appropriate distance measure
is very crucial; however, the selection criteria still depends largely
on the nature of data itself, especially in time series data. Though
the Euclidean distance is commonly used to measure the dissimilarity
between two time series, it has been shown that DTW distance is more
appropriate and produces more accurate results. Sakoe-Chiba Band (S-C
Band) \cite{108244} originally speeds up the DTW calculation and
later has been introduced to be used as a DTW global constraint. In
addition, the S-C Band was first implemented for the speech community,
and the width of the global constraint was fixed to be 10\% of time
series length. However, recent work \cite{RatanamahatanaK05} reveals
that the classification accuracy depends solely on this global constraint;
the size of the constraint depends on the properties of the data at
hands. To determine a suitable size, all possible widths of the global
constraint are tested, and the band with the maximum training accuracy
is selected. 

Ratanamahatana-Keogh Band (R-K Band) \cite{RatanamahatanaK04} has
been introduced to generalize the global constraint model represented
by a one-dimensional array. The size of the array and the maximum
constraint value is limited to the length of the time series. And
the main feature of the R-K band is the multi bands, where each band
is representing each class of data. Unlike the single S-C band, this
multi R-K bands can be adjusted as needed according to its own class\textquoteright{}
warping path. 

Although the R-K band allows great flexibility to adjust the global
constraint, a learning algorithm is needed to discover the \textquoteleft{}best\textquoteright{}
multi R-K bands. In the original work of R-K Band, a hill climbing
search algorithm with two heuristic functions (accuracy and distance
metrics) is proposed. The search algorithm climbs though a space by
trying to increase/decrease specific parts of the bands until terminal
conditions are met. However, this learning algorithm still suffers
from an \textquoteleft{}overfitting\textquoteright{} phenomenon since
an accuracy metric is used as a heuristic function to guide the search.

To solve this problem, we propose two new learning algorithms, i.e.,
band boundary extraction and iterative learning. The band boundary
extraction method first obtains a maximum, mean, and mode of the path\textquoteright{}s
positions on the DTW distance matrix, and the iterative learning,
band's structures are adjusted in each round of the iteration to a
Silhouette Index \cite{rousseeuw1987sga}. We run both algorithms
and the band that gives better results. In prediction step, the 1-NN
using Dynamic Time Warping distance with this discovered band is used
to classify unlabeled data. Note that a lower bound, LB\_Keogh \cite{KeoghR05},
is also used to speed up our 1-NN classification.

The rest of this paper is organized as follows. Section 2 gives some
important background for our proposed work. In Section 3, we introduce
our approach, the two novel learning algorithms. Section 4 contains
an experimental evaluation including some examples of each dataset.
Finally, we conclude this paper in Section 5.

\section{Background}

Our novel learning algorithms are based on four major fundamental
concepts, i.e., Dynamic Time Warping (DTW) distance, Sakoe-Chiba band
(S-C band), Ratanamahatana-Keogh band (R-K band), and Silhouette index,
which are briefly described in the following sections.

\subsection{Dynamic Time Warping Distance}

Dynamic Time Warping (DTW) \cite{RatanamahatanaK05} distance is a
well-known similarity measure based on shape. It uses a dynamic programming
technique to find all possible warping paths, and selects the one
with the minimum distance between two time series. To calculate the
distance, it first creates a distance matrix, where each element in
the matrix is a cumulative distance of the minimum of three surrounding
neighbors. Suppose we have two time series, a sequence $Q$ of length
$n$ ($Q=q_{1},q_{2},\ldots,q_{i},\ldots,q_{n}$) and a sequence $C$
of length $m$ ($C=c_{1},c_{2},\ldots,c_{j},\ldots,c_{m}$). First,
we create an $n$-by-$m$ matrix, where every ($i,j$) element of
the matrix is the cumulative distance of the distance at ($i,j$)
and the minimum of three neighboring elements, where $1\leq i\leq n$
and $1\leq j\leq m$. We can define the ($i,j$) element, $\gamma_{i,j}$,
of the matrix as:

\begin{equation}
\gamma_{i,j}=d_{i,j}+\min\left\{ \gamma_{i-1,j},\gamma_{i,j-1},\gamma_{i-1,j-1}\right\} \label{eq:dtw}\end{equation}

\noindent where $d_{i,j}=\left(c_{i}-q_{j}\right)^{2}$ is the squared
distance of $q_{i}$ and $c_{j}$, and $\gamma_{i,j}$ is the summation
of $d_{i,j}$ and the the minimum cumulative distance of three elements
surrounding the ($i,j$) element. Then, to find an optimal path, we
choose the path that yields a minimum cumulative distance at ($n,m$),
which is defined as: 

\begin{equation}
D_{DTW}(Q,C)=\underset{\forall w\in P}{\min}\left\{ \sqrt{\overset{K}{\underset{k=1}{\sum}}d_{w_{k}}}\right.\label{eq:dtw2}\end{equation}

\noindent where $P$ is a set of all possible warping paths, $w_{k}$
is ($i,j$) at $k^{\mathrm{th}}$ element of a warping path, and $K$
is the length of the warping path.

\begin{figure}
\noindent \begin{centering}
\begin{tabular}{cc}
\includegraphics[width=3.5cm]{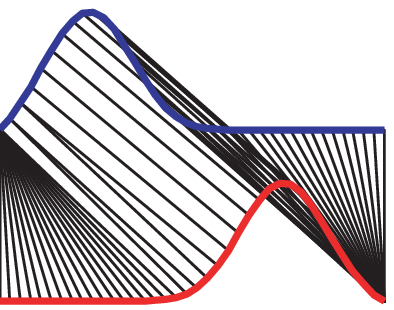} & \includegraphics[width=3.5cm]{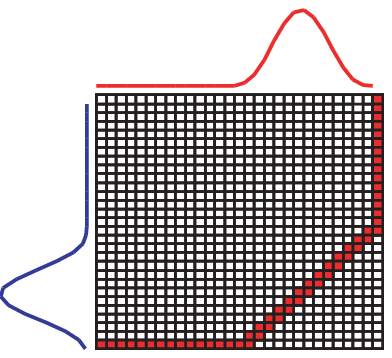}\tabularnewline
\end{tabular}
\par\end{centering}

\caption{DTW without using global constraint may introduce an unwanted warping.}
\label{Flo:dtw1}
\end{figure}

\begin{figure}
\noindent \begin{centering}
\begin{tabular}{cc}
\includegraphics[width=3.5cm]{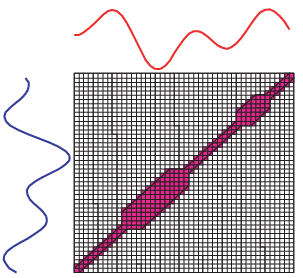} & \includegraphics[width=3.5cm]{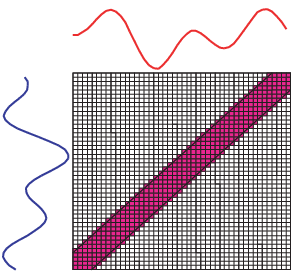}\tabularnewline
(a) & (b)\tabularnewline
\multicolumn{2}{c}{\includegraphics[width=3.5cm]{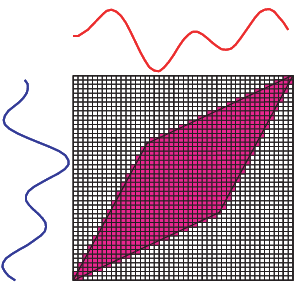}}\tabularnewline
\multicolumn{2}{c}{(c)}\tabularnewline
\end{tabular}
\par\end{centering}

\caption{Global constraint examples of (a) R-K band (b) S-C band, and (c) Itakura
Parallelogram.}
\label{Flo:dtw2}
\end{figure}

In reality, DTW may not give the best mapping according to our need
because it will try its best to find the minimum distance. It may
generate the unwanted path. For example, in Figure \ref{Flo:dtw1}
\cite{RatanamahatanaK05}, without global constraint, DTW will find
its optimal mapping between the two time series. However, in many
cases, this is probably not what we intend, when the two time series
are expected to be of different classes. We can resolve this problem
by limiting the permissible warping paths using a global constraint.
Two well-known global constraints, Sakoe-Chiba band and Itakura Parallelogram
\cite{1162641}, and a recent representation, Ratanamahatana-Keogh
band (R-K band), have been proposed, Figure \ref{Flo:dtw2} \cite{RatanamahatanaK04}
shows an example for each type of the constraints. %

\subsection{Sakoe-Chiba Band}

Sakoe-Chiba band (S-C band), shown in Figure \ref{Flo:dtw2} (b),
is one of the simplest and most popular global constraints, originally
introduced to be used for speech community. The width of this global
constraint is generally set to be 10\% of the time series length.
However, recent work \cite{RatanamahatanaK05} has shown that the
different sizes of the band can be used towards a more accurate classification.
We therefore need to test all possible widths of the global constraint
so that the best width could be discovered. An evaluation function
is needed to justify the selection. We commonly use accuracy metric
(a training accuracy) as a measurement. Table \ref{Flo:scband} shows
an algorithm in finding the best warping window for S-C band by decreasing
the band size by 1\% in each step. This function receives a set of
data $T$ as an input, and gives the best warping window (\emph{best\_band})
as an output. Note that if an evaluation value is equal to the best
evaluation value, we prefer the smaller warping window size. %
\begin{table}
\caption{Finding the best warping window.}

\noindent \begin{centering}
\begin{tabular}{cl}
\multicolumn{2}{l}{\noun{\scriptsize Function}{\scriptsize{} {[}}\emph{\scriptsize best\_band}{\scriptsize {]}
= }\noun{\scriptsize BestWarping}{\scriptsize{} {[}$T${]}}}\tabularnewline
\midrule
{\scriptsize 1} & \emph{\scriptsize best\_evaluate}{\scriptsize{} = }\noun{\scriptsize NegativeInfinite}{\scriptsize ;}\tabularnewline
{\scriptsize 2} & {\scriptsize for ($k$ = 100 to 0)}\tabularnewline
{\scriptsize 3} & {\scriptsize ~~~~~}\emph{\scriptsize $bandk$}{\scriptsize{} =
S-C band at $k$\% width;}\tabularnewline
{\scriptsize 4} & {\scriptsize ~~~~~}\emph{\scriptsize $evaluate$}{\scriptsize{}
= evaluate($band_{k}$);}\tabularnewline
{\scriptsize 5} & {\scriptsize ~~~~~if (}\emph{\scriptsize $evaluate$}{\scriptsize{}
>= }\emph{\scriptsize best\_evaluate}{\scriptsize )}\tabularnewline
{\scriptsize 6} & {\scriptsize ~~~~~~~~~~}\emph{\scriptsize best\_evaluate}{\scriptsize{}
= }\emph{\scriptsize evaluate}{\scriptsize ;}\tabularnewline
{\scriptsize 7} & {\scriptsize ~~~~~~~~~~}\emph{\scriptsize best\_band}{\scriptsize{}
= $band_{k}$}\tabularnewline
{\scriptsize 8} & {\scriptsize ~~~~~endif}\tabularnewline
{\scriptsize 9} & {\scriptsize endfor}\tabularnewline
\bottomrule
\end{tabular}
\par\end{centering}

\label{Flo:scband}
\end{table}

\subsection{Ratanamahatana-Keogh Band}

Ratanamahatana-Keogh band (R-K band) \cite{RatanamahatanaK04} is
a general model of a global constraint specified by a one-dimensional
array $R$, i.e., $R=r_{1},r_{2},\ldots,r_{i},\ldots,r_{n}$ where
$n$ is the length of time series, and $r_{i}$ is the height above
the diagonal in $y$ direction and the width to the right of the diagonal
in $x$ direction. Each $r_{i}$ value is arbitrary, therefore R-K
band is also an arbitrary-shape global constraint, as shown in Figure
\ref{Flo:dtw2} (a). Note that when $r_{i}=0$, where $1\leq i\leq n$,
this R-K band represents the Euclidean distance, and when $r_{i}=n$,
where $1\leq i\leq n$, this R-K band represents the original DTW
distance with no global constraint. The R-K band is also able to represent
the S-C band by giving all $r_{i}=c$, where $c$ is the width of
a global constraint. Moreover, the R-K band is a multi band model
which can be effectively used to represent one band for each class
of data. This flexibility is a great advantage; however, the higher
the number of classes, the higher the time complexity, as we have
to search through such a large space. 

Since determining the optimal R-K band for each training set is highly
computationally intensive, a hill climbing and heuristic functions
have been introduced to guide which part of space should be evaluated.
A space is defined as a segment of a band to be increased or decreased.
In the original work, two heuristic functions, accuracy metric and
distance metric, are used to evaluate a state. The accuracy metric
is evaluated from the training accuracy using leaving-one-out 1-NN,
and the distance metric is a ratio of the mean DTW distances of correctly
classified and incorrectly classified objects. However, these heuristic
functions do not reflect the true quality of a band because empirically,
we have found that the resulting bands tend to \textquoteleft{}overfit\textquoteright{}
the training data. 

Two searching directions are considered, i.e., forward search, and
backward search. In forward search, we start from the Euclidean distance
(all $r_{i}$ in $R$ equal to 0), and parts of the band are gradually
increased in each searching step. In the case where two bands have
the same heuristic value, the wider band is selected. On the other
hand, in backward search, we start from a very large band (all $r_{i}$
in $R$ equal to $n$, where $n$ is the length of time series), and
parts of the band are gradually decreased in each searching step.
If two bands have the same heuristic value, the tighter band is chosen.%
\begin{table}[t]
\caption{The pseudo code for multiple R-K bands learning.}

\noindent \begin{centering}
\begin{tabular}{cl}
\multicolumn{2}{l}{\noun{\scriptsize Function}{\scriptsize{} {[}$band${]} = }\noun{\scriptsize Learning}{\scriptsize {[}$T$,$threshold${]}}}\tabularnewline
\midrule
{\scriptsize 1} & \emph{\scriptsize $N$}{\scriptsize = size of }\emph{\scriptsize $T$}{\scriptsize ;}\tabularnewline
{\scriptsize 2} & \emph{\scriptsize $L$}{\scriptsize = length of data in $T$;}\tabularnewline
{\scriptsize 3} & {\scriptsize initialize $band_{i}$ for $i$ = 1 to $c$;}\tabularnewline
{\scriptsize 4} & {\scriptsize foreachclass $i$ = 1 to $c$}\tabularnewline
{\scriptsize 5} & {\scriptsize ~~~enqueue(1, $L$, $Queue_{i}$);}\tabularnewline
{\scriptsize 6} & {\scriptsize endfor}\tabularnewline
{\scriptsize 7} & \emph{\scriptsize best\_evaluate}{\scriptsize{} = evaluate($T$, $band$);}\tabularnewline
{\scriptsize 8} & {\scriptsize while !empty($Queue$)}\tabularnewline
{\scriptsize 9} & {\scriptsize ~~~foreachclass $i$ = 1 to $c$}\tabularnewline
{\scriptsize 10} & {\scriptsize ~~~~~~if !empty($Queue_{i}$)}\tabularnewline
{\scriptsize 11} & {\scriptsize ~~~~~~~~~{[}$start$, $end${]} = dequeue($Queue_{i}$)}\tabularnewline
{\scriptsize 12} & {\scriptsize ~~~~~~~~~$adjusta$ble = adjust($bandi$, $start$,
$end$);}\tabularnewline
{\scriptsize 13} & {\scriptsize ~~~~~~~~~if $adjustable$}\tabularnewline
{\scriptsize 14} & {\scriptsize ~~~~~~~~~~~~$evaluate$= evaluate($T$, $band$);}\tabularnewline
{\scriptsize 15} & {\scriptsize ~~~~~~~~~~~~if $evaluate$ > }\emph{\scriptsize best\_evaluate}\tabularnewline
{\scriptsize 16} & {\scriptsize ~~~~~~~~~~~~~~~}\emph{\scriptsize best\_evaluate}{\scriptsize{}
= $evaluate$;}\tabularnewline
{\scriptsize 17} & {\scriptsize ~~~~~~~~~~~~~~~enqueue($start$, $end$,
$Queue_{i}$);}\tabularnewline
{\scriptsize 18} & {\scriptsize ~~~~~~~~~~~~else}\tabularnewline
{\scriptsize 19} & {\scriptsize ~~~~~~~~~~~~~~~undo\_adjustment($band_{i}$,
$start$, $end$);}\tabularnewline
{\scriptsize 20} & {\scriptsize ~~~~~~~~~~~~~~~if ($start$ \textendash{}
$end$) / 2 \ensuremath{\ge} $threshold$}\tabularnewline
{\scriptsize 21} & {\scriptsize ~~~~~~~~~~~~~~~~~~enqueue($start$,
$mid$-1, $Queue_{i}$);}\tabularnewline
{\scriptsize 22} & {\scriptsize ~~~~~~~~~~~~~~~~~~enqueue($mid$, $end$,
$Queue_{i}$);}\tabularnewline
{\scriptsize 23} & {\scriptsize ~~~~~~~~~~~~~~~endif}\tabularnewline
{\scriptsize 24} & {\scriptsize ~~~~~~~~~~~~endif}\tabularnewline
{\scriptsize 25} & {\scriptsize ~~~~~~~~~endif}\tabularnewline
{\scriptsize 26} & {\scriptsize ~~~~~~endif}\tabularnewline
{\scriptsize 27} & {\scriptsize ~~~endfor}\tabularnewline
{\scriptsize 28} & {\scriptsize endwhile}\tabularnewline
\bottomrule
\end{tabular}
\par\end{centering}

\label{Flo:rkband2}
\end{table}
\begin{figure}
\noindent \begin{centering}
\includegraphics[width=7cm]{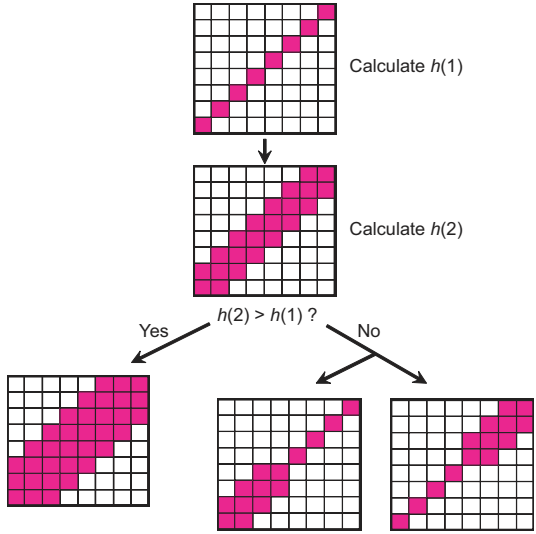}
\par\end{centering}

\caption{An illustration of the concept in R-K band forward searching algorithm. }
\label{Flo:rkband}
\end{figure}

Our learning algorithm starts from first enqueuing the starting- and
ending-parts of the R-K Band. In each iteration, these values are
dequeued, and used as a boundary for a band increase/decrease. And
then the adjusted band is evaluated. If the heuristic value is higher
than the current best heuristic value, the same start and end values
are enqueued. If not, this part is further divided into two equal
subparts before being enqueued, as shown in Figure \ref{Flo:rkband}
\cite{RatanamahatanaK04}. The iterations are continued until a termination
condition is met. Table 2 shows the pseudo code for this multiple
R-K bands learning.

\subsection{Silhouette Index}

Silhouette index (SI) \cite{rousseeuw1987sga} or Silhouette width
is a well-known clustering validity technique, originally used to
determine a number of clusters in a dataset. This index measures the
\textquoteleft{}quality\textquoteright{} of separation and compactness
of a clustered dataset, so the number of cluster is determined by
selecting the number that gives maximum index value. 

The Silhouette index is based on a compactness-separation measurement
which consists of an inter-cluster distance (a distance between two
different-cluster data) and an intra-cluster distance (a distance
between two same-cluster data). A good clustered dataset means that
the dataset has high inter-cluster distance and low intra-cluster
distance. In other words, a good clustered dataset is the dataset
that different-cluster are well separated, and the same-cluster data
are well grouped together. The Silhouette index for each data $i$
is defined by the following equation: 

\begin{equation}
s(i)=\frac{b(i)-a(i)}{\max\left\{ b(i),a(i)\right\} }\label{eq:sil1}\end{equation}

\begin{equation}
b(i)=\underset{c\in C\wedge c\neq label(i)}{\min}\left(\frac{1}{N_{D_{c}}}\underset{j\in D_{c}}{\sum}d(i,j)\right)\label{eq:sil2}\end{equation}

\begin{equation}
a(i)=\frac{1}{N_{D_{label(i)}}}\underset{j\in D_{label(i)}\wedge i\neq j}{\sum}d(i,j)\label{eq:sil3}\end{equation}

\noindent where $s(i)$ is the Silhouette index of $i^{\mathrm{th}}$
data, $b(i)$ is the minimum average distance between the $i^{\mathrm{th}}$
data and each of the different-cluster data, and $a(i)$ is the average
distance between the $i^{\mathrm{th}}$ data and each of the same-cluster
data. In Equations (\ref{eq:sil2}) and (\ref{eq:sil3}), $C$ is
a set of all possible clusters, $D_{c}$ is a set of data in cluster
$c$, $N_{D_{c}}$ is the size of $D_{c}$, and $d$($i,j$) is the
distance measure function comparing between $i^{\mathrm{th}}$ and
$j^{\mathrm{th}}$ data. Note that the $s(i)$ ranges between -1 and
1. Having $s(i)$ close to 1 means that data is well separated. Global
Silhouette index (GS) for a dataset is calculated as follows.

\begin{equation}
GS=\frac{1}{c}\overset{c}{\underset{j=1}{\sum}S_{j}}\label{eq:sil4}\end{equation}

\begin{equation}
S_{c}=\frac{1}{M}\overset{M}{\underset{i=1}{\sum}}s(i)\label{eq:sil5}\end{equation}

\noindent where $c$ is the number of clusters, $S_{c}$ is the Silhouette
index for cluster $c$, and $M$ is the number of data in cluster
$c$. The pseudo code for the Silhouette index function is shown in
Table \ref{Flo:sil1}. %
\begin{table}
\caption{Silhouette index function.}

\noindent \begin{centering}
\begin{tabular}{cl}
\multicolumn{2}{l}{\noun{\scriptsize Function}{\scriptsize{} {[}$index${]} = }\noun{\scriptsize Silhouette}{\scriptsize {[}$D${]}}}\tabularnewline
\midrule
{\scriptsize 1} & {\scriptsize $N$= size of $D$;}\tabularnewline
{\scriptsize 2} & \emph{\scriptsize sum\_All}{\scriptsize{} = 0;}\tabularnewline
{\scriptsize 3} & {\scriptsize foreachclass $j$ = 1 to $c$}\tabularnewline
{\scriptsize 4} & {\scriptsize ~~~~~$M$ = size of $Dj$;}\tabularnewline
{\scriptsize 5} & {\scriptsize ~~~~~}\emph{\scriptsize sum\_Class}{\scriptsize{}
= 0;}\tabularnewline
{\scriptsize 6} & {\scriptsize ~~~~~for $i$ = 1 to $M$}\tabularnewline
{\scriptsize 7} & {\scriptsize ~~~~~~~~~~$b$ = b($i$);}\tabularnewline
{\scriptsize 8} & {\scriptsize ~~~~~~~~~~$a$ = a($i$);}\tabularnewline
{\scriptsize 9} & {\scriptsize ~~~~~~~~~~$s$ = ($b$ \textendash{} $a$)
/ max($b$, $a$);}\tabularnewline
{\scriptsize 10} & {\scriptsize ~~~~~~~~~~}\emph{\scriptsize sum\_Class}{\scriptsize{}
+= $s$;}\tabularnewline
{\scriptsize 11} & {\scriptsize ~~~~~endfor}\tabularnewline
{\scriptsize 12} & {\scriptsize ~~~~~}\emph{\scriptsize sum\_All}{\scriptsize{} +=
}\emph{\scriptsize sum\_Class}{\scriptsize{} / $M$;}\tabularnewline
{\scriptsize 13} & {\scriptsize endfor}\tabularnewline
{\scriptsize 14} & {\scriptsize $index$= }\emph{\scriptsize sum\_All}{\scriptsize{} /
$c$;}\tabularnewline
\bottomrule
\end{tabular}
\par\end{centering}

\label{Flo:sil1}
\end{table}

\section{Methodology}

In this section, we describe our approach, developed from the techniques
described in Section 2, i.e., the DTW distance, the best warping window
for Sakoe-Chiba band, multiple R-K bands, and the Silhouette index.
In brief, our approach consists of 5 major parts: 1) data preprocessing
that reduces the length of time series data, 2) our proposed band
boundary extraction algorithm, 3) finding the best warping window
for Sakoe-Chiba band, 4) our proposed iterative R-K band learning,
and 5) prediction for unlabeled data.

Our approach requires three input parameters, i.e., a set of training
data $T$, a set of unlabeled data (test data) $P$, the maximum complexity
that depends on time and computational resources, and the bound of
a warping window size. In data preprocessing step, we could reduce
the computational complexity in case of very long time series data
using interpolation function, both in training and test data. After
that, we try to find the best R-K band by running the band boundary
extraction algorithm. The best warping window is calculated and is
used as an initial band of our proposed iterative learning. After
learning have finished, two R-K bands are compared and the better
one is selected. Finally, we calculate a training accuracy and make
predictions for the test data using 1-NN with the DTW distance and
the best band, enhanced with LB\_Keogh lower bound to speed up our
classification approach. The prediction result A along with the training
accuracy are returned as shown in Table \ref{Flo:our}. %
\begin{table}
\caption{Our classification approach.}

\noindent \begin{centering}
\begin{tabular}{cl}
\multicolumn{2}{l}{\noun{\scriptsize Function}{\scriptsize{} {[}$A$, $accuracy${]} =
}\noun{\scriptsize OurApproach}{\scriptsize {[}$T$, $P$, $complexity$,
$bound${]}}}\tabularnewline
\midrule
{\scriptsize 1} & {\scriptsize $L$= length of data in $T$;}\tabularnewline
{\scriptsize 2} & {\scriptsize {[}$T$, $L$, $P${]} = preprocess($T$, $P$, $L$,
$complexity$);}\tabularnewline
{\scriptsize 3} & {\scriptsize {[}}\emph{\scriptsize best\_band}{\scriptsize , }\emph{\scriptsize best\_heuristic}{\scriptsize {]}
= band\_extraction($T$);}\tabularnewline
{\scriptsize 4} & {\scriptsize $R$= best\_warping($T$, $bound$);}\tabularnewline
{\scriptsize 5} & {\scriptsize {[}$band$, $heuristic${]} = iterative\_learning($T$,
$R$, $L$, $bound$);}\tabularnewline
{\scriptsize 6} & {\scriptsize if ($heuristic$ > }\emph{\scriptsize best\_heuristic}{\scriptsize )}\tabularnewline
{\scriptsize 7} & {\scriptsize ~~~~~}\emph{\scriptsize best\_heuristic}{\scriptsize{}
= }\emph{\scriptsize $heuristic$}{\scriptsize ;}\tabularnewline
{\scriptsize 8} & {\scriptsize ~~~~~}\emph{\scriptsize best\_band}{\scriptsize{}
= $band$;}\tabularnewline
{\scriptsize 9} & {\scriptsize endif}\tabularnewline
{\scriptsize 10} & {\scriptsize $accuracy$= leave\_one\_out($T$, }\emph{\scriptsize best\_band}{\scriptsize );}\tabularnewline
{\scriptsize 11} & {\scriptsize $A$= predict($T$, $P$, }\emph{\scriptsize best\_band}{\scriptsize );}\tabularnewline
\bottomrule
\end{tabular}
\par\end{centering}

\label{Flo:our}
\end{table}

\subsection{Data Preprocessing}

Since the classification prediction time may because a major constraint,
a data preprocessing step is needed. In this step, we approximate
the calculation complexity and try to reduce the complexity exceeding
the threshold. Our approximated complexity is mainly based on the
number of items in the training data, its length, and the number of
heuristic function evaluations. Suppose we have $n$ training data
with $m$ data points in length, we can calculate the complexity by
the following equation.

\[
complexity(n,m)=\log(n^{2}\times m^{2})\]

\noindent where a logarithm function is added to bring down the value
to a more manageable range for users.

To decrease the complexity, we could reduce the length of each individual
time series by using typical interpolation function. The new length
of time series is set to be the current length divided by two. We
keep reducing the time series length until the complexity is smaller
than the user\textquoteright{}s defined threshold. Table \ref{Flo:pre}
shows the preprocessing steps on a set of training data $T$, a set
of unlabeled data $P$, the original length $L$, and the complexity
threshold. In this work, we set this threshold value to 9, according
to resources and the time constraint for this 24-hour Workshop and
Challenge on Time Series Classification. %
\begin{table}
\caption{Data preprocessing step.}

\noindent \begin{centering}
\begin{tabular}{cl}
\multicolumn{2}{l}{\noun{\scriptsize Function}{\scriptsize{} {[}$NewT$, $NewP$, $NewL${]}
= }\noun{\scriptsize PreProcess}{\scriptsize {[}$T$, $P$, $L$,
$threshold${]}}}\tabularnewline
\midrule
{\scriptsize 1} & {\scriptsize $alpha$= complexity($T$, $L$);}\tabularnewline
{\scriptsize 2} & {\scriptsize set $NewT$ = $T$, $NewP$ = $P$, and $NewL$ = $L$;}\tabularnewline
{\scriptsize 3} & {\scriptsize while ($alpha$ > $threshold$)}\tabularnewline
{\scriptsize 4} & {\scriptsize ~~~~~$NewL$ = $NewL$ / 2;}\tabularnewline
{\scriptsize 5} & {\scriptsize ~~~~~$NewT$ = interpolate($NewT$, $NewL$);}\tabularnewline
{\scriptsize 6} & {\scriptsize ~~~~~$NewP$ = interpolate($NewP$, $NewL$);}\tabularnewline
{\scriptsize 7} & {\scriptsize ~~~~~$alpha$ = complexity($NewT$, $NewL$);}\tabularnewline
{\scriptsize 8} & {\scriptsize endwhile}\tabularnewline
\bottomrule
\end{tabular}
\par\end{centering}

\label{Flo:pre}
\end{table}
\begin{table}
\caption{Boundary Band Extraction Algorithm.}

\noindent \centering{}\begin{tabular}{cl}
\multicolumn{2}{l}{\noun{\scriptsize Function}{\scriptsize{} {[}}\emph{\scriptsize best\_band}{\scriptsize ,
}\emph{\scriptsize best\_heuristic}{\scriptsize {]} = }\noun{\scriptsize BandExtraction}{\scriptsize {[}$T${]}}}\tabularnewline
\midrule
{\scriptsize 1} & {\scriptsize $N$= size of $T$;}\tabularnewline
{\scriptsize 2} & {\scriptsize $L$= length of data in $T$;}\tabularnewline
{\scriptsize 3} & {\scriptsize initialize }\emph{\scriptsize path\_matrix}{\scriptsize{}
= new array {[}$L${]}{[}$L${]};}\tabularnewline
{\scriptsize 4} & {\scriptsize initialize $R$ for $Max$, $Mean$, and $Mode$}\tabularnewline
{\scriptsize 5} & {\scriptsize foreachclass ($k$ = 1 to $c$)}\tabularnewline
{\scriptsize 6} & {\scriptsize ~~~~~$Nk$ = size of $Tk$;}\tabularnewline
{\scriptsize 7} & {\scriptsize ~~~~~for ($i$ = 1 to $N$)}\tabularnewline
{\scriptsize 8} & {\scriptsize ~~~~~~~~~~for ($j$ = 1 to $N$)}\tabularnewline
{\scriptsize 9} & {\scriptsize ~~~~~~~~~~~~~~~if ($i$ != $j$)}\tabularnewline
{\scriptsize 10} & {\scriptsize ~~~~~~~~~~~~~~~~~~~~$Path$ = dtw\_path($i$,
$j$);}\tabularnewline
{\scriptsize 11} & {\scriptsize ~~~~~~~~~~~~~~~~~~~~for (all point
$p$ in $Path$)}\tabularnewline
{\scriptsize 12} & {\scriptsize ~~~~~~~~~~~~~~~~~~~~~~~~~}\emph{\scriptsize path\_matrix}{\scriptsize {[}$p.x${]}{[}$p.y${]}++;}\tabularnewline
{\scriptsize 13} & {\scriptsize ~~~~~~~~~~~~~~~~~~~~endfor}\tabularnewline
{\scriptsize 14} & {\scriptsize ~~~~~~~~~~~~~~~endif}\tabularnewline
{\scriptsize 15} & {\scriptsize ~~~~~~~~~~endfor}\tabularnewline
{\scriptsize 16} & {\scriptsize ~~~~~endfor}\tabularnewline
{\scriptsize 17} & {\scriptsize ~~~~~for (i = 0 to $L$)}\tabularnewline
{\scriptsize 18} & {\scriptsize ~~~~~~~~~~$Max_{k}${[}$i${]} = maximum warping
path at $r_{i}$}\tabularnewline
{\scriptsize 19} & {\scriptsize ~~~~~~~~~~$Mean_{k}${[}$i${]} = mean warping
path at $r_{i}$}\tabularnewline
{\scriptsize 20} & {\scriptsize ~~~~~~~~~~$Mode_{k}${[}$i${]} = mode warping
path at $r_{i}$}\tabularnewline
{\scriptsize 21} & {\scriptsize ~~~~~endfor}\tabularnewline
{\scriptsize 22} & {\scriptsize end}\tabularnewline
{\scriptsize 23} & {\scriptsize {[}}\emph{\scriptsize best\_band}{\scriptsize , }\emph{\scriptsize best\_heuristic}{\scriptsize {]}
= bestband($Max$, $Mean$, $Mode$);}\tabularnewline
\bottomrule
\end{tabular}\label{Flo:bb2}
\end{table}
\begin{figure}[b]
\noindent \begin{centering}
\begin{tabular}{ccc}
\includegraphics[width=2cm]{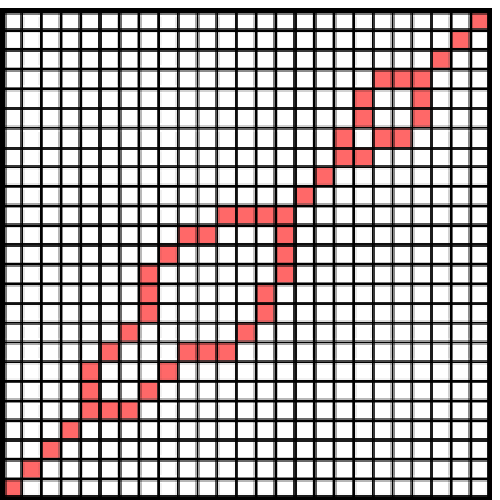} & \includegraphics[width=2cm]{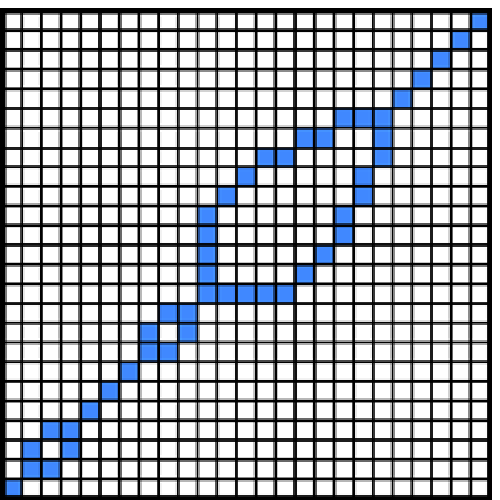} & \includegraphics[width=2cm]{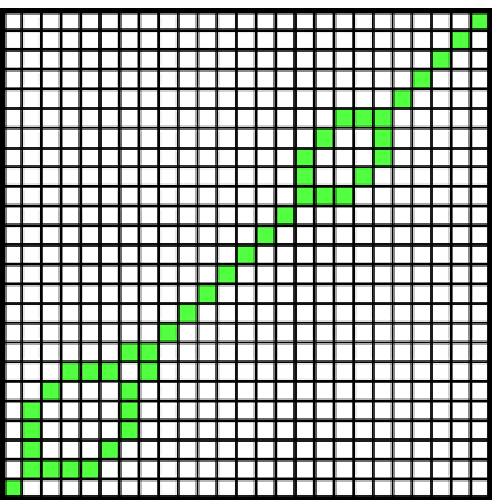}\tabularnewline
 & (a) & \tabularnewline
\includegraphics[width=2cm]{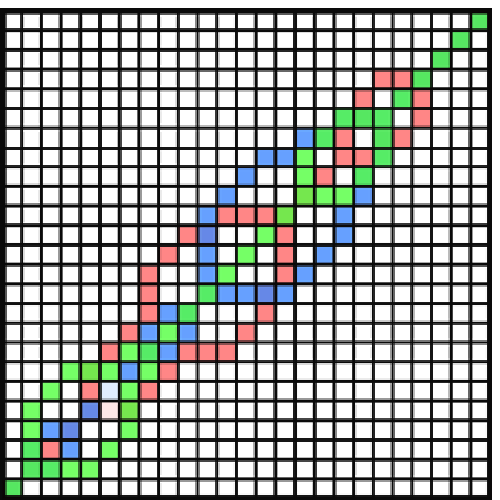} &  & \includegraphics[width=2cm]{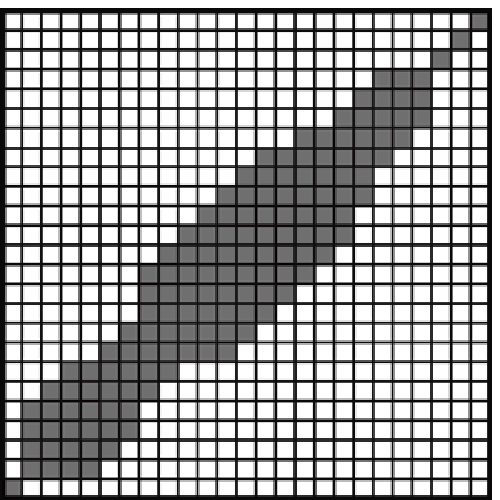}\tabularnewline
(b) &  & (c)\tabularnewline
\end{tabular}
\par\end{centering}

\caption{The illustration of MaxBand calculation. (a) finding all possible
warping paths, (b) plotting all paths in a path matrix, and (c) calculating
maximum value for each $r_{i}$}
\label{Flo:bb1}
\end{figure}

\subsection{Boundary Band Extraction}

Since the multi R-K band model allows a learning algorithm to create
a different band for each different class. This boundary band extraction
algorithm is derived from a simple intuition for each of the same-class
data, we first calculate all their DTW distances, save all the warping
paths, and plot those paths on a matrix (called a path matrix). After
that, we will determine an appropriate R-K band. For each $r_{i}$
of this R-K band, the $r_{i}$ value is set to be the maximum between
height above the diagonal in $y$ direction and width to the right
of the diagonal in $x$ direction in the path matrix. We repeat these
steps to every possible class in the dataset; we call this R-K band
a $MaxBand$. Similarly, the band extraction is performed using mean
average and mode instead of the maximum, resulting in a $MeanBand$
and a $ModeBand$, respectively. Figure \ref{Flo:bb1} illustrates
the steps in creating a $MaxBand$. From these calculations, three
multiple R-K bands are generated. The evaluation function is used
to select the best band to be returned as an output of this algorithm.
Table \ref{Flo:bb2} shows the band boundary extraction algorithm
on a set of training data $T$ and return the best R-K band and the
best heuristic value. %
\begin{table}
\caption{A search algorithm for the best S-C band warping window.}

\noindent \centering{}\begin{tabular}{cl}
\multicolumn{2}{l}{\noun{\scriptsize Function}{\scriptsize{} {[}$R${]} = B}\noun{\scriptsize estWarping}{\scriptsize {[}$T$,$bound${]}}}\tabularnewline
\midrule
{\scriptsize 1} & {\scriptsize $N$= size of $T$;}\tabularnewline
{\scriptsize 2} & \emph{\scriptsize best\_heuristic}{\scriptsize{} = }\noun{\scriptsize NegativeInfinite}{\scriptsize ;}\tabularnewline
{\scriptsize 3} & {\scriptsize for ($k$ = $bound$ to 0)}\tabularnewline
{\scriptsize 4} & {\scriptsize ~~~~~$bandk$= S-C Band at $k$\% width;}\tabularnewline
{\scriptsize 5} & {\scriptsize ~~~~~$heuristic$= evaluate($bandk$);}\tabularnewline
{\scriptsize 6} & {\scriptsize ~~~~~if ($heuristic$ > }\emph{\scriptsize best\_heuristic}{\scriptsize )}\tabularnewline
{\scriptsize 7} & {\scriptsize ~~~~~~~~~~}\emph{\scriptsize best\_heuristic}{\scriptsize{}
= $heuristic$;}\tabularnewline
{\scriptsize 8} & {\scriptsize ~~~~~~~~~~$R$ = $k$;}\tabularnewline
{\scriptsize 9} & {\scriptsize ~~~~~endif}\tabularnewline
{\scriptsize 10} & {\scriptsize endfor}\tabularnewline
\bottomrule
\end{tabular}\label{Flo:bw}
\end{table}

\subsection{Finding the Best Warping Window}

In this step, we try to achieve the best warping window of Sakoe-Chiba
band to be an input of our proposed iterative R-K band learning. This
function is slightly different from the original one in that we bound
the maximum width of the warping window and we use our evaluation
function (heuristic function) instead of the typical training accuracy.
A simple pseudo code is described in Table \ref{Flo:bw} below. A
set of training data $T$ and a maximum warping window size are required
in discovering the best warping window $R$. %
\begin{table}
\caption{Our proposed Iterative R-K band learning algorithm.}

\noindent \centering{}\begin{tabular}{cl}
\multicolumn{2}{l}{\noun{\scriptsize Function}{\scriptsize{} {[}$band${]} = }\noun{\scriptsize IterativeLearning}{\scriptsize {[}$T$,
$R$, $L$, $bound${]}}}\tabularnewline
\midrule
{\scriptsize 1} & {\scriptsize initialize }\emph{\scriptsize best\_band}{\scriptsize $_{i}$
for }\emph{\scriptsize i }{\scriptsize = 0 to $c$ equals to }\emph{\scriptsize R}{\scriptsize \%
of }\emph{\scriptsize L}\tabularnewline
{\scriptsize 2} & {\scriptsize threshold = }\emph{\scriptsize L}{\scriptsize{} / 2;}\tabularnewline
{\scriptsize 3} & \emph{\scriptsize best\_heuristic}{\scriptsize{} = evaluate(}\emph{\scriptsize T}{\scriptsize ,
}\emph{\scriptsize band}{\scriptsize );}\tabularnewline
{\scriptsize 4} & {\scriptsize while (}\emph{\scriptsize threshold}{\scriptsize ) <
1}\tabularnewline
{\scriptsize 5} & {\scriptsize ~~~~~}\emph{\scriptsize fw\_band}{\scriptsize{} =
forward\_learning(}\emph{\scriptsize T}{\scriptsize , $L$, $band$,
$bound$);}\tabularnewline
{\scriptsize 6} & {\scriptsize ~~~~~}\emph{\scriptsize bw\_band}{\scriptsize{} =
backward\_learning(}\emph{\scriptsize $T$}{\scriptsize , $L$, $band$,
$bound$);}\tabularnewline
{\scriptsize 7} & {\scriptsize ~~~~~}\emph{\scriptsize fw\_heuristic}{\scriptsize{}
= evaluate($T$, }\emph{\scriptsize fw\_band}{\scriptsize );}\tabularnewline
{\scriptsize 8} & {\scriptsize ~~~~~}\emph{\scriptsize bw\_heuristic}{\scriptsize{}
= evaluate($T$, }\emph{\scriptsize bw\_band}{\scriptsize );}\tabularnewline
{\scriptsize 9} & {\scriptsize ~~~~~$band$ = maximum heuristic value band;}\tabularnewline
{\scriptsize 10} & {\scriptsize ~~~~~$heuristic$ = maximum heuristic value;}\tabularnewline
{\scriptsize 11} & {\scriptsize ~~~~~if $heuristic$ = }\emph{\scriptsize best\_heuristic}\tabularnewline
{\scriptsize 12} & {\scriptsize ~~~~~~~~~~$threshold$ = $threshold$ / 2;}\tabularnewline
{\scriptsize 13} & {\scriptsize ~~~~~endif}\tabularnewline
{\scriptsize 14} & {\scriptsize endwhile}\tabularnewline
\bottomrule
\end{tabular}\label{Flo:il1}
\end{table}

\subsection{Iterative Band Learning}

The iterative R-K band learning is extended from the original learning
that it will repeat the learning again and again until a heuristic
value no longer increases. In the first step, we initialize all the
multi R-K bands with $R$\% Sakoe-Chiba band, where $R$ is the output
from the best finding warping window algorithm. We also set a learning
threshold to be half of the time series length, and the initial bands
are evaluated

In each iteration, our proposed algorithm learns a new R-K band starting
with the previous R-K band learning result both in forward and backward
direction. We also run both forward and backward learning and select
the best band which gives a higher heuristic value. If the heuristic
value is the same as the best heuristic value, the threshold is divided
by two; otherwise we update the best heuristic value. We repeat these
steps until the threshold falls below 1. Table \ref{Flo:il1} shows
our proposed algorithm, iterative R-K band learning, which requires
a set of training data $T$, a best warping window $R$, the length
of time series $L$, and the bound of warping window.%
\begin{table}
\caption{Our proposed R-K band learning algorithm.}

\noindent \centering{}\begin{tabular}{cl}
\multicolumn{2}{l}{\noun{\scriptsize Function}{\scriptsize{} {[}$R${]} = }\noun{\scriptsize ProposedLearning}{\scriptsize {[}$T$,
$threshold$, $band$, $bound${]}}}\tabularnewline
\midrule
{\scriptsize 1} & {\scriptsize $L$= length of data in T;}\tabularnewline
{\scriptsize 2} & {\scriptsize foreachclass i = 1 to c}\tabularnewline
{\scriptsize 3} & {\scriptsize enqueue(1, L, i, Queue);}\tabularnewline
{\scriptsize 4} & {\scriptsize endfor}\tabularnewline
{\scriptsize 5} & {\scriptsize best\_heuristic = evaluate(T, band);}\tabularnewline
{\scriptsize 6} & {\scriptsize while !empty(Queue)}\tabularnewline
{\scriptsize 7} & {\scriptsize ~~~{[}start, end, label{]} = randomly\_dequeue(Queue)}\tabularnewline
{\scriptsize 8} & {\scriptsize ~~~adjustable = adjust(bandlabel, start, end, bound);}\tabularnewline
{\scriptsize 9} & {\scriptsize ~~~if adjustable}\tabularnewline
{\scriptsize 10} & {\scriptsize ~~~~~~heuristic = evaluate(T, band);}\tabularnewline
{\scriptsize 11} & {\scriptsize ~~~~~~if heuristic > best\_heuristic}\tabularnewline
{\scriptsize 12} & {\scriptsize ~~~~~~~~~best\_heuristic = heuristic;}\tabularnewline
{\scriptsize 13} & {\scriptsize ~~~~~~~~~enqueue(start, end, label, Queue);}\tabularnewline
{\scriptsize 14} & {\scriptsize ~~~~~~else}\tabularnewline
{\scriptsize 15} & {\scriptsize ~~~~~~~~~undo\_adjustment(bandlabel, start,
end);}\tabularnewline
{\scriptsize 16} & {\scriptsize ~~~~~~~~~if (start \textendash{} end) / 2 \ensuremath{\ge}
threshold}\tabularnewline
{\scriptsize 17} & {\scriptsize ~~~~~~~~~~~~enqueue(start, mid-1, label,
Queue);}\tabularnewline
{\scriptsize 18} & {\scriptsize ~~~~~~~~~~~~enqueue(mid, end, label, Queue);}\tabularnewline
{\scriptsize 19} & {\scriptsize ~~~~~~~~~endif}\tabularnewline
{\scriptsize 20} & {\scriptsize ~~~~~~endif}\tabularnewline
{\scriptsize 21} & {\scriptsize ~~~endif}\tabularnewline
{\scriptsize 22} & {\scriptsize endwhile}\tabularnewline
\bottomrule
\end{tabular}\label{Flo:il2}
\end{table}

We have modified the original multi R-K bands learning by changing
its data structure. We replace multi queues, which are originally
assigned for each class by only one single queue with an addition
parameter label to each start-end object. This new queue will draw
an object randomly instead of last-in-first-out (LIFO) manner. In
addition, we also change an adjustment function by adding a new parameter
bound to limit forward learning not to increase the band\textquoteright{}s
size exceeding limited bound. Table \ref{Flo:il2} shows the proposed
R-K band learning on a set of training data $T$, a learning threshold,
an initial band, and the bound of warping window.

\subsection{Evaluation Function}

From Section 2.4, we have briefly described the utility and the algorithm
of the Silhouette index. This index is commonly used to measure the
quality of a clustered dataset; however, we can utilize this Silhouette
index as a heuristic function to measure the quality of a distance
measure as well. The DTW distance with multi R-K bands is a distance
measure that requires one additional parameter, $Band$, specifying
the R-K band to be used (since the multi R-K bands contain one band
for each class). Table \ref{Flo:ef} shows the evaluation function
derived from the original Silhouette index. 

\begin{equation}
s(i,Band)=\frac{b(i,Band)-a(i,Band)}{\max\left\{ b(i,Band),a(i,Band)\right\} }\label{eq:eva1}\end{equation}

\begin{equation}
b(i,Band)=\underset{c\in C\wedge c\neq label(i)}{\min}\left(\frac{1}{N_{D_{c}}}\underset{j\in D_{c}}{\sum}d(i,j,Band_{label(j)})\right)\label{eq:eva2}\end{equation}

\begin{equation}
a(i,Band)=\frac{1}{N_{D_{c}}}\underset{j\in D_{label(j)}\wedge i\neq j}{\sum}d(i,j,Band_{label(j)})\label{eq:eva3}\end{equation}

\noindent %
\begin{table}
\caption{An evaluation (heuristic) function.}

\noindent \centering{}\begin{tabular}{cl}
\multicolumn{2}{l}{\noun{\scriptsize Function}{\scriptsize{} {[}$index${]} = }\noun{\scriptsize Evaluate}{\scriptsize {[}$D$,$B${]}}}\tabularnewline
\midrule
{\scriptsize 1} & {\scriptsize $N$= size of $D$;}\tabularnewline
{\scriptsize 2} & \emph{\scriptsize sum\_All}{\scriptsize{} = 0;}\tabularnewline
{\scriptsize 3} & {\scriptsize foreachclass ($j$ = 1 to $c$)}\tabularnewline
{\scriptsize 4} & {\scriptsize ~~~~~$M$ = size of $D_{j}$;}\tabularnewline
{\scriptsize 5} & {\scriptsize ~~~~~}\emph{\scriptsize sum\_Class}{\scriptsize{}
= 0;}\tabularnewline
{\scriptsize 6} & {\scriptsize ~~~~~for ($i$ = 1 to $M$)}\tabularnewline
{\scriptsize 7} & {\scriptsize ~~~~~~~~~~$b$ = b($i$, $B$);}\tabularnewline
{\scriptsize 8} & {\scriptsize ~~~~~~~~~~$a$ = a($i$, $B$);}\tabularnewline
{\scriptsize 9} & {\scriptsize ~~~~~~~~~~$s$ = ($b$ \textendash{} $a$)
/ max($b$, $a$);}\tabularnewline
{\scriptsize 10} & {\scriptsize ~~~~~~~~~~}\emph{\scriptsize sum\_Class}{\scriptsize{}
+= $s$;}\tabularnewline
{\scriptsize 11} & {\scriptsize ~~~~~endfor}\tabularnewline
{\scriptsize 12} & {\scriptsize ~~~~~}\emph{\scriptsize sum\_All}{\scriptsize{} +=
}\emph{\scriptsize sum\_Class}{\scriptsize{} / $M$;}\tabularnewline
{\scriptsize 13} & {\scriptsize endfor}\tabularnewline
{\scriptsize 14} & {\scriptsize $index$= }\emph{\scriptsize sum\_All}{\scriptsize{} /
$c$;}\tabularnewline
\bottomrule
\end{tabular}\label{Flo:ef}
\end{table}

\subsection{Data Prediction}

After the best multi R-K bands are discovered, we use 1-Nearest Neighbor
as a classifier and the Dynamic Time Warping distance measure with
these best R-K bands for prediction in the test data to predict a
set of unlabeled data. The LB\_Keogh lower bound is also used to speed
up the DTW computation.

\section{Experimental Evaluation}

To evaluate the performance, we use our approach, described in Section
3, to classify all 20 contest datasets, and then send our predicted
results and the expected accuracies to the contest organizer. The
results are calculated by the contest organizer and subsequently sent
back to us.

\subsection{Datasets}

\begin{figure}
\noindent \begin{centering}
\begin{tabular}{cc}
\includegraphics{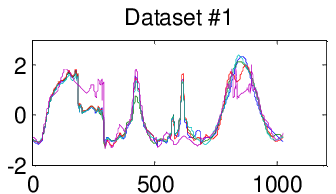} & \includegraphics{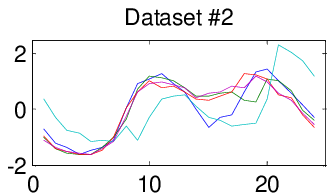}\tabularnewline
\includegraphics{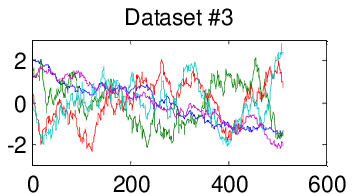} & \includegraphics{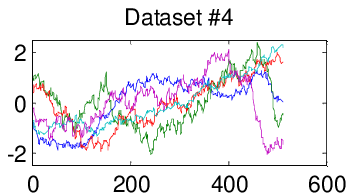}\tabularnewline
\includegraphics{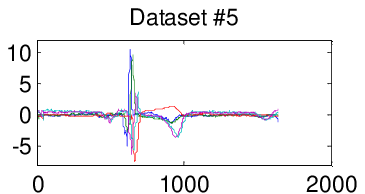} & \includegraphics{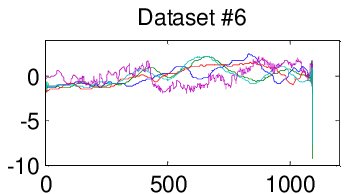}\tabularnewline
\includegraphics{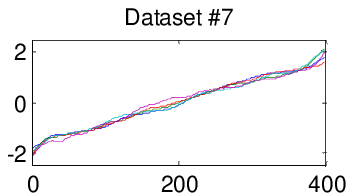} & \includegraphics{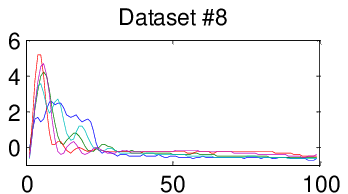}\tabularnewline
\includegraphics{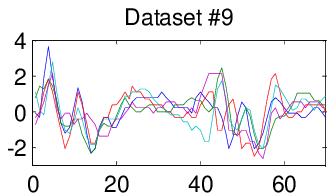} & \includegraphics{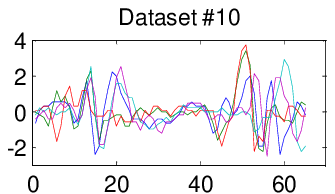}\tabularnewline
\includegraphics{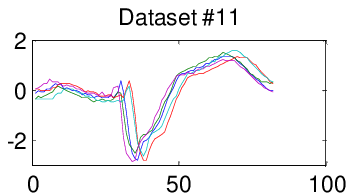} & \includegraphics{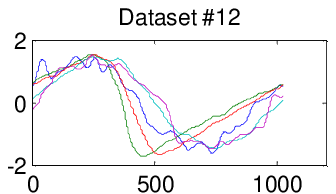}\tabularnewline
\includegraphics{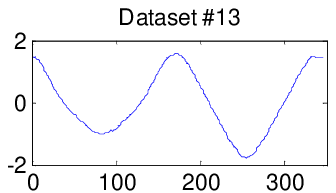} & \includegraphics{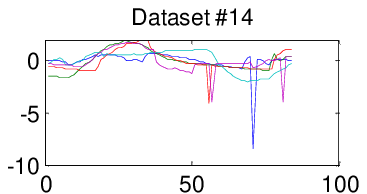}\tabularnewline
\includegraphics{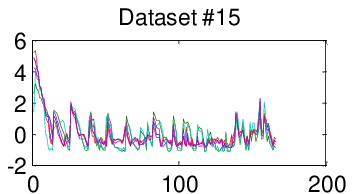} & \includegraphics{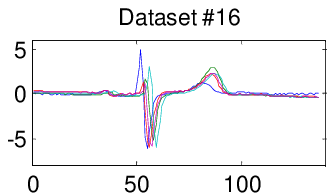}\tabularnewline
\includegraphics{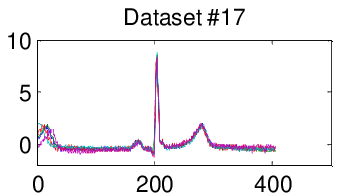} & \includegraphics{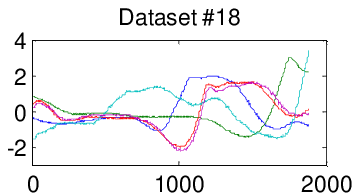}\tabularnewline
\includegraphics{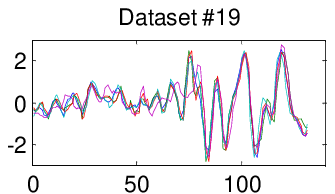} & \includegraphics{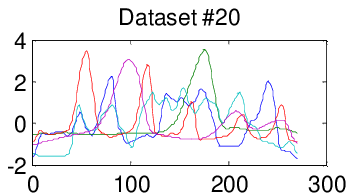}\tabularnewline
\end{tabular}
\par\end{centering}

\caption{Some samples from each of the 20 datasets.}
\label{Flo:dataset1}
\end{figure}

We use the datasets from the Workshop and Challenge on Time Series
Classification, held in conjunction with the thirteenth SIGKDD 2007
conference. The datasets are from very diverse domains (e.g., stock
data, medical data, etc.); some are from real-world problems, and
some are synthetically generated. The amount of training data and
its length in each dataset also vary from the size of 20 to 1000 training
instances and the length of 30 to 2000 data points. In addition, all
data are individually normalized using Z-normalization. Examples of
each dataset are shown in Figure \ref{Flo:dataset1}, and the datasets\textquoteright{}
properties are shown in Table \ref{Flo:dataset2}. %
\begin{table}
\caption{The dataset properties.}

\noindent \centering{}\begin{tabular}{|c|c|c|c|c|}
\hline 
\textbf{\scriptsize Dataset} & \textbf{\scriptsize \#Classes} & \textbf{\scriptsize Training} & \textbf{\scriptsize Test} & \textbf{\scriptsize Length of each}\tabularnewline
 &  & \textbf{\scriptsize data size} & \textbf{\scriptsize data size} & \textbf{\scriptsize time series}\tabularnewline
\hline
\hline 
{\scriptsize 1} & {\scriptsize 8} & {\scriptsize 55} & {\scriptsize 2345} & {\scriptsize 1024}\tabularnewline
\hline 
{\scriptsize 2} & {\scriptsize 2} & {\scriptsize 67} & {\scriptsize 1029} & {\scriptsize 24}\tabularnewline
\hline 
{\scriptsize 3} & {\scriptsize 2} & {\scriptsize 367} & {\scriptsize 1620} & {\scriptsize 512}\tabularnewline
\hline 
{\scriptsize 4} & {\scriptsize 2} & {\scriptsize 178} & {\scriptsize 1085} & {\scriptsize 512}\tabularnewline
\hline 
{\scriptsize 5} & {\scriptsize 4} & {\scriptsize 40} & {\scriptsize 1380} & {\scriptsize 1639}\tabularnewline
\hline 
{\scriptsize 6} & {\scriptsize 5} & {\scriptsize 155} & {\scriptsize 308} & {\scriptsize 1092}\tabularnewline
\hline 
{\scriptsize 7} & {\scriptsize 6} & {\scriptsize 25} & {\scriptsize 995} & {\scriptsize 398}\tabularnewline
\hline 
{\scriptsize 8} & {\scriptsize 10} & {\scriptsize 381} & {\scriptsize 760} & {\scriptsize 99}\tabularnewline
\hline 
{\scriptsize 9} & {\scriptsize 2} & {\scriptsize 20} & {\scriptsize 601} & {\scriptsize 70}\tabularnewline
\hline 
{\scriptsize 10} & {\scriptsize 2} & {\scriptsize 27} & {\scriptsize 953} & {\scriptsize 65}\tabularnewline
\hline 
{\scriptsize 11} & {\scriptsize 2} & {\scriptsize 23} & {\scriptsize 1139} & {\scriptsize 82}\tabularnewline
\hline 
{\scriptsize 12} & {\scriptsize 3} & {\scriptsize 1000} & {\scriptsize 8236} & {\scriptsize 1024}\tabularnewline
\hline 
{\scriptsize 13} & {\scriptsize 4} & {\scriptsize 16} & {\scriptsize 306} & {\scriptsize 345}\tabularnewline
\hline 
{\scriptsize 14} & {\scriptsize 2} & {\scriptsize 20} & {\scriptsize 1252} & {\scriptsize 84}\tabularnewline
\hline 
{\scriptsize 15} & {\scriptsize 3} & {\scriptsize 467} & {\scriptsize 3840} & {\scriptsize 166}\tabularnewline
\hline 
{\scriptsize 16} & {\scriptsize 2} & {\scriptsize 23} & {\scriptsize 861} & {\scriptsize 136}\tabularnewline
\hline 
{\scriptsize 17} & {\scriptsize 2} & {\scriptsize 73} & {\scriptsize 936} & {\scriptsize 405}\tabularnewline
\hline 
{\scriptsize 18} & {\scriptsize 7} & {\scriptsize 100} & {\scriptsize 550} & {\scriptsize 1882}\tabularnewline
\hline 
{\scriptsize 19} & {\scriptsize 12} & {\scriptsize 200} & {\scriptsize 2050} & {\scriptsize 131}\tabularnewline
\hline 
{\scriptsize 20} & {\scriptsize 15} & {\scriptsize 267} & {\scriptsize 638} & {\scriptsize 270}\tabularnewline
\hline
\end{tabular}\label{Flo:dataset2}
\end{table}

\subsection{Results}

The predicted result is generated after running our algorithm to find
the best R-K band within the competition\textquoteright{}s 24-hour
time constraint. More specifically, the predicted accuracy is calculated
by computing leaving-one-out cross validation on the training dataset.
Table \ref{Flo:result} shows our predicted accuracies and testing
accuracies for all 20 datasets which are calculated and are returned
to the contest organizer. Because of the small number of training
data, the predicted accuracy and the test accuracy are different in
some cases.%
\begin{table}
\caption{The predicted and test accuracies.}

\noindent \centering{}\begin{tabular}{|c|c|c|}
\hline 
\textbf{\scriptsize Dataset} & \textbf{\scriptsize Predicted accuracy} & \textbf{\scriptsize Test accuracy}\tabularnewline
\hline
\hline 
{\scriptsize 1} & {\scriptsize 0.9636} & {\scriptsize 0.6505}\tabularnewline
\hline 
{\scriptsize 2} & {\scriptsize 0.9403} & {\scriptsize 0.9161}\tabularnewline
\hline 
{\scriptsize 3} & {\scriptsize 0.4714} & {\scriptsize 0.3491}\tabularnewline
\hline 
{\scriptsize 4} & {\scriptsize 0.9494} & {\scriptsize 0.9231}\tabularnewline
\hline 
{\scriptsize 5} & {\scriptsize 0.9500} & {\scriptsize 0.8537}\tabularnewline
\hline 
{\scriptsize 6} & {\scriptsize 0.1871} & {\scriptsize 0.6099}\tabularnewline
\hline 
{\scriptsize 7} & {\scriptsize 1.0000} & {\scriptsize 0.8714}\tabularnewline
\hline 
{\scriptsize 8} & {\scriptsize 0.7428} & {\scriptsize 0.9346}\tabularnewline
\hline 
{\scriptsize 9} & {\scriptsize 0.9500} & {\scriptsize 0.8488}\tabularnewline
\hline 
{\scriptsize 10} & {\scriptsize 0.8519} & {\scriptsize 0.8507}\tabularnewline
\hline 
{\scriptsize 11} & {\scriptsize 0.9565} & {\scriptsize 0.8489}\tabularnewline
\hline 
{\scriptsize 12} & {\scriptsize 0.8570} & {\scriptsize 0.8353}\tabularnewline
\hline 
{\scriptsize 13} & {\scriptsize 0.9375} & {\scriptsize 0.7250}\tabularnewline
\hline 
{\scriptsize 14} & {\scriptsize 0.9000} & {\scriptsize 0.9276}\tabularnewline
\hline 
{\scriptsize 15} & {\scriptsize 0.6017} & {\scriptsize 0.4435}\tabularnewline
\hline 
{\scriptsize 16} & {\scriptsize 0.7391} & {\scriptsize 0.8645}\tabularnewline
\hline 
{\scriptsize 17} & {\scriptsize 0.8904} & {\scriptsize 0.9346}\tabularnewline
\hline 
{\scriptsize 18} & {\scriptsize 0.2900} & {\scriptsize 0.5049}\tabularnewline
\hline 
{\scriptsize 19} & {\scriptsize 0.9500} & {\scriptsize 0.9660}\tabularnewline
\hline 
{\scriptsize 20} & {\scriptsize 0.6966} & {\scriptsize 0.9275}\tabularnewline
\hline
\end{tabular}\label{Flo:result}
\end{table}

\section{Conclusion}

In this work, we propose a new efficient time series classification
algorithm based on 1-Nearest Neighbor classification using the Dynamic
Time Warping distance with multi R-K bands as a global constraint.
To select the best R-K band, we use our two proposed learning algorithms,
i.e., band boundary extraction algorithm and iterative learning. Silhouette
index is used as a heuristic function for selecting the band that
yields the best prediction accuracy. The LB\_Keogh lower bound is
also used in data prediction step to speed up the computation.

\section{Acknowledgments}

We would like to thank the Scientific PArallel Computer Engineering
(SPACE) Laboratory, Chulalongkorn University for providing a cluster
we have used in this contest.

\bibliographystyle{plain}

\end{document}